\documentclass[runningheads]{llncs}

\usepackage{eccv}

\usepackage{eccvabbrv}
\usepackage{colortbl} 
\usepackage{xcolor}   
\usepackage{mathtools}
\usepackage{color}
\usepackage{dsfont}
\usepackage{graphicx}
\usepackage{booktabs}
\usepackage{multicol}
\usepackage{multirow}
\usepackage{algorithm2e}
\usepackage{xcolor}
\usepackage{enumitem}
\usepackage{makecell}
\usepackage{colortbl}
\usepackage[normalem]{ulem}

\usepackage{graphicx}
\usepackage{booktabs}

\usepackage[colorinlistoftodos,prependcaption,textsize=tiny]{todonotes}
\presetkeys{todonotes}{inline}{}
\usepackage{comment}

\usepackage[accsupp]{axessibility}  

\newcommand{\dcfaceOGNosp}{DCFace$_{eg}$}
\newcommand{\dcfaceOG}{\dcfaceOGNosp~}
\newcommand{\dcfaceOGANosp}{DCFace$_{ega}$}
\newcommand{\dcfaceOGA}{\dcfaceOGANosp~}

\newcommand{\dcfaceAllNosp}{DCFace$_{all}$}
\newcommand{\dcfaceAll}{\dcfaceAllNosp~}

\newcommand{\dcfaceNosp}{DCFace}
\newcommand{\dcface}{\dcfaceNosp~}

\newcommand{\digifaceNosp}{DigiFace}
\newcommand{\digiface}{\digifaceNosp~}

\newcommand{\casiaNosp}{CASIA}
\newcommand{\casia}{\casiaNosp~}

\newcommand{\lfwcNosp}{LFW-C}
\newcommand{\lfwc}{\lfwcNosp~}

\newcommand{\lfwNosp}{LFW}
\newcommand{\lfw}{\lfwNosp~}

\newcommand{\calfwNosp}{CALFW}
\newcommand{\calfw}{\calfwNosp~}

\newcommand{\agedbNosp}{AgeDB}

\newcommand{\cplfwNosp}{CPLFW}
\newcommand{\cplfw}{\cplfwNosp~}

\newcommand{\cfpfpNosp}{CFP-FP}

\newcommand{\favcidNosp}{FAVCI2D}
\newcommand{\favcid}{\favcidNosp~}

\newcommand{\genderNosp}{\textbf{\textit{g}}}
\newcommand{\originNosp}{\textbf{\textit{e}}}
\newcommand{\ageNosp}{\textbf{\textit{a}}}
\newcommand{\poseNosp}{\textbf{\textit{p}}}

\newcommand{\gender}{\genderNosp~}
\newcommand{\origin}{\originNosp~}
\newcommand{\age}{\ageNosp~}
\newcommand{\pose}{\poseNosp~}

\newcommand{\ffaceNosp}{\textit{ff}}

\newcommand{\fface}{\ffaceNosp}

\usepackage[pagebackref,breaklinks,colorlinks]{hyperref}

\usepackage{orcidlink}

\begin{document}

\title{Toward Fairer Face Recognition Datasets}

\titlerunning{Fairer Face Datasets}

\author{Alexandre Fournier-Montgieux$^*$\inst{1}\orcidlink{0009-0002-7744-3179} \and
Michael Soumm$^*$\inst{1}\orcidlink{0009-0009-0435-9903} \and
Adrian Popescu\inst{1}\orcidlink{0000-0002-8099-824X} \and
Bertrand Luvison\inst{1}\orcidlink{0000-0003-2475-3712} \and
Hervé Le Borgne\inst{1}\orcidlink{0000-0003-0520-8436}
}

\authorrunning{A.~Fournier-Montgieux et al.}

\institute{Université Paris-Saclay, CEA, List, F-91120, Palaiseau, France
}

\maketitle

\begin{abstract}

Face recognition and verification are two computer vision tasks whose performance has progressed with the introduction of deep representations.  
However, ethical, legal, and technical challenges due to the sensitive character of face data and biases in real training datasets hinder their development.
Generative AI addresses privacy by creating fictitious identities, but fairness problems persist. 
We promote fairness by introducing a demographic attributes balancing mechanism in generated training datasets.
We experiment with an existing real dataset, three generated training datasets, and the balanced versions of a diffusion-based dataset.
We propose a comprehensive evaluation that considers accuracy and fairness equally and includes a rigorous regression-based statistical analysis of attributes.
The analysis shows that balancing reduces demographic unfairness.
Also, a performance gap persists despite generation becoming more accurate with time. 
The proposed balancing method and comprehensive verification evaluation promote fairer and transparent face recognition and verification.

\end{abstract}

\def\thefootnote{*}\footnotetext{These authors contributed equally to this work}\def\thefootnote{\arabic{footnote}}
\section{Introduction}
\label{sec_intro}

The development of AI recently raised numerous questions about its role in society.
The debate is vivid regarding automatic face recognition and face verification technologies (FRT and FVT) due to the sensitive nature of inferences made with these technologies~\cite{van2020ethical}. 
FRT aims to recognize a person among different identities, while FVT assesses whether two face images correspond to the same identity.
A socially acceptable development of these technologies should consider legal, ethical, and technical factors.
Legal initiatives such as the EU's General Data Protection Regulation (GDPR) are worthy attempts to regulate such sensitive AI technologies, but they do not apply globally.
Civil society projects performed ethical auditing of face recognition and notably highlighted data collection and usage problems~\cite{Exposing_ai}.
They built public pressure contributing to the withdrawal of most publicly-available large-scale face recognition datasets, such as MS-Celeb-1M~\cite{cao2018vggface2}, MegaFace~\cite{kemelmacher2016megaface}, and VGGFace2~\cite{guo2016ms}.
Although these withdrawal are justified, they limit the availability of training datasets and negatively affects the open FRT and FRV development and their programmatic auditing.
Following a more general trend~\cite{ho2022cascaded}, data generation has gained traction in face recognition~\cite{bae2023digiface,d2024improving,kim2023dcface,qiu2021synface}. 
This approach is appealing since training datasets include fictitious identities, and privacy problems are consequently reduced.
Technical auditing highlighted the negative impact of biases related to attributes such as gender, age, and ethnicity in real~\cite{albiero2021gendered,buolamwini2018gender,karkkainen2019fairface,popescu2022face,sarridis2023towards,van2020ethical} and, recently, generated datasets~\cite{perera2023analyzing}.
However, they do not address fairness during the training datasets collection and structuring.

\begin{figure}[tb]
    \centering
    \includegraphics[width=\linewidth]{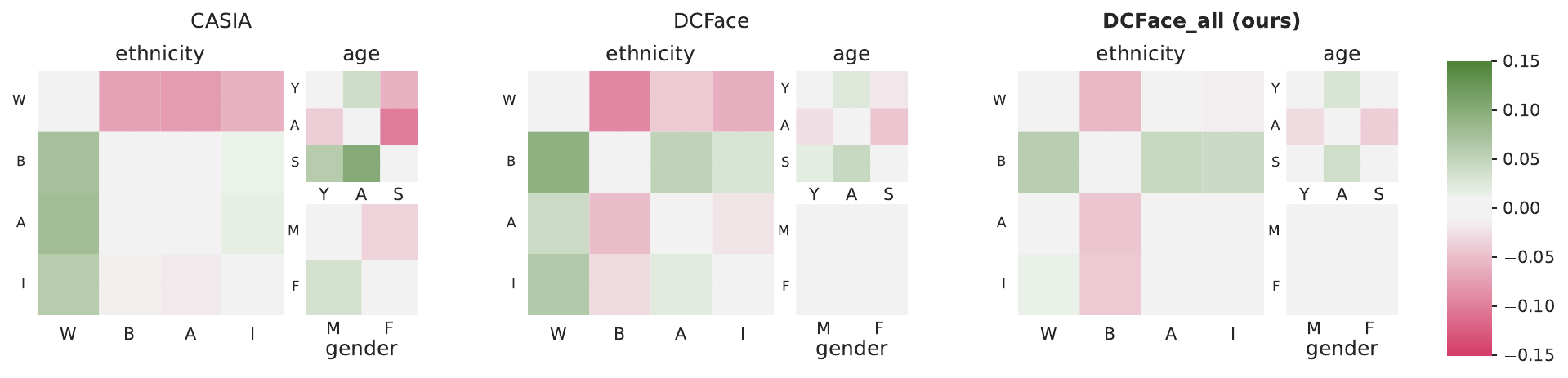}
    \caption{Fairness of the three training datasets in terms of \textit{equal opportunity}~\cite{hardt2016equality_opportunity} when considering the \textit{advantaged outcome} as being person differenciation: for each combination $(i,j)$ of an attribute, we consider the difference in opportunity between the demographics $i$ and $j$. Values close to 0 indicate an equality of outcome. Considered demographics : \textit{White, Black, Asian, Indian}; \textit{Young, Adult, Senior}; \textit{Male, Female}}. 
    \label{fig_teaser}
\end{figure}
Informed by existing works, we take a more comprehensive approach to address fairness during the entire face recognition and verification pipeline. 

We find that insufficient importance is given to fairness, even in recent works~\cite{bae2023digiface,kim2023dcface}, despite the extensive availability of bias-related literature~\cite{albiero2021gendered,buolamwini2018gender,karkkainen2019fairface,popescu2022face,sarridis2023towards,van2020ethical}.
For instance, in~\autoref{fig_teaser}, our analysis shows that \textit{Black} people have a lower \textit{equal opportunity}~\cite{hardt2016equality_opportunity} than \textit{White} people, which is also the case for \textit{Asian} and \textit{Indian}, both for the real dataset CASIA~\cite{yi2014learning} and the synthetic one DCface~\cite{kim2023dcface}.
To address these issues, we propose the following contributions:
\begin{itemize}[noitemsep,topsep=0pt,leftmargin=*]
\item Evidencing biases in existing training datasets. In particular, we show unfairness persistence in synthetic datasets, particularly for recent diffusion-based models~\cite{kim2023dcface}.
\item Introducing a simple yet effective fairness promoting mechanism in a diffusion-based training dataset creation, that can be incorporated into any generative pipeline, that does not affect accuracy of further FVT approaches. 
\item Promoting systematic use of fairness besides accuracy in face verification evaluation. Including fairness leads to a comprehensive understanding of the merits and limitations of compared approaches. 
\item Proposing a rigorous statistical analysis framework that reveals the roles of individual attributes.
\end{itemize}
Code and associated resources are released to facilitate the adoption of fairness in FRT and FRV: \url{https://github.com/afm215/TowardFairerFaceRecognitionSets}.

\section{Related Work}
\label{sec_related}
\noindent\textbf{Training dataset structure} determines how deep representations are trained. 
The authors of~\cite{schroff2015facenet} rely on a dataset including tens of millions of unique identities and a small number of samples per identity, with a contrastive loss.
However, the collection of such datasets is practically difficult, and most works~\cite{conti2022mitigating,deng2019arcface,kim2022adaface,zhao2018towards} use datasets with fewer identities and more samples per identity~\cite{cao2018vggface2,guo2016ms,kemelmacher2016megaface,yi2014learning}. 
They adapt the loss function by adding task-specific components such as angular margin loss~\cite{deng2019arcface}, image quality~\cite{kim2022adaface}, or intra-class variance~\cite{conti2022mitigating}.

\noindent\textbf{Real training datasets} are usually created by scrapping a large number of images from publicly available sources~\cite{kemelmacher2016megaface,schroff2015facenet} then cleaning them~\cite{cao2018vggface2,guo2016ms,yi2014learning} to reduce the number of unrepresentative samples.
First, subjects' consent is impossible to obtain at scale at this poses a serious legal challenge when collecting sensitive data such as identified faces. 
Licensing is a second important challenge, with most datasets~\cite{cao2018vggface2,guo2016ms,yi2014learning} including copyrighted photos. 
The lawfulness of distributing copyrighted content is a longstanding discussion that applies to other computer vision tasks~\cite{quang2021does}. 
It was revived by the success of foundation models trained with very large datasets~\cite{scao2022bloom}.
Third, existing large datasets exhibit demographic (gender, ethnicity, age)~\cite{popescu2022face,sarridis2023towards,wang2019racial}, face characteristics (size, make-up, hairstyle)~\cite{albiero2020does,albiero2021gendered,terhorst2021comprehensive} and visual biases~\cite{zhao2018towards}.
They are mostly a reflection of the sampling bias affecting visual datasets~\cite{fabbrizzi2022bias_visual_datasets}. 
These biases affect underrepresented segments~\cite{buolamwini2018gender,karkkainen2019fairface,sarridis2023towards} and should be addressed to improve fairness.
Taken together, these problems make the sustainable publication of real datasets very complicated, as proven by the withdrawal of most resources~\cite{cao2018vggface2,guo2016ms,kemelmacher2016megaface} following public pressure~\cite{van2020ethical}.

\noindent\textbf{Synthetic datasets} have the potential to reduce or remove privacy, copyright, and unfairness compared to real datasets~\cite{kim2023dcface}. 
Computer graphics techniques are used in~\cite{bae2023digiface,wood2021fake} to render diversified face images, and strong augmentations are added to increase accuracy. 
Most works rely on generative AI, with~\cite{zhao2017dual} being an early example that uses dual-agent GANs to generate photorealistic faces.
The authors of~\cite{qiu2021synface} identify the lack of variability of generated images as a central challenge and propose identity and domain mixup to improve synthetic datasets. 
Diffusion models were used very recently~\cite{kim2023dcface} to create identities and to diversify their samples based on a style bank.  
Synthetic datasets have the advantage of including fictitious identities, a characteristic that alleviates the privacy and copyright issues associated with real FRT datasets. 
However, problems remain regarding data replication in GANs~\cite{feng2021gans} and diffusion models~\cite{somepalli2023diffusion}, posing a privacy problem when the generation process is not controlled for privacy. 
A  solution that obfuscates the identity while preserving the attributes of the face for GANs was introduced in~\cite{barattin2023attribute}.
Generative models may improve fairness because the dataset design is fully controlled. 
However, when uncontrolled, synthetic datasets reproduce and even exacerbate the biases of real datasets in a constrained evaluation setting~\cite{perera2023analyzing}.
While interesting, this approach is not directly applicable to diffusion models, which produce higher quality synthetic datasets~\cite{kim2023dcface}.
Compared to~\cite{perera2023analyzing}, we perform a more thorough and realistic evaluation of fairness for real and synthetic datasets, both existing and proposed. 

\noindent\textbf{Face verification} is a classical yet still open research topic.
Given a pair of images, the task is to determine whether they belong to the same identity. 
A threshold separates the positive and negative pairs~\cite{robinson2020face}.
More efforts are needed to mitigate biases and integrate fairness in the verification evaluation process.
Demographic attributes balance deserves particular attention because it is required for analyzing potentially serious discrimination~\cite{sarridis2023towards}. 
Bias mitigation during verification relies on adapting the process to demographic segments.
The authors of~\cite{robinson2020face,terhorst2020comparison} propose adaptive threshold-based approaches to improve fairness.
Ethnicity-related bias is addressed by learning disparate margins per demographic segment in the representation space~\cite{yang2021ramface,wang2021meta,wang2020mitigating} or by suppressing attribute-related information in the model~\cite{sarridis2023flac}. 
While technically interesting, these metrics are ethically and legally problematic in practice since they assume disparate treatment of human subjects by AI-based systems. 

Fairness evaluation can be improved by designing demographically-diversified verification datasets~\cite{grother2019face,popescu2022face,wang2019racial} and integrating rich demographic metadata in them~\cite{sarridis2023towards}.
Performance saturation on datasets such as \lfwNosp~\cite{LFWTechUpdate} is also problematic.
It is mainly due to the random choice of negative pairs and, following~\cite{phillips2012good,popescu2022face,wang2019racial}, we advocate for selecting hard negative images to make verification more realistic. 

\section{Compared Face Recognition Training Datasets}
Works such as~\cite{barattin2023attribute,buolamwini2018gender,popescu2022face} audits face recognition and verification to quantify the biases appearing after deployment. 
Other works focus on fairness for particular attributes such as gender~\cite{albiero2020does} or ethnic origin~\cite{wang2019racial}.
Informed by these studies, we hypothesize that training dataset biases are mainly an effect of an uncontrolled or weakly controlled dataset creation process. 
Our main objective is to improve fairness in FRT and FRV while maintaining accuracy.
We first present the fairness promotion approach applied to synthetic FRT training datasets. 
Second, we discuss the essential characteristics of the compared FRT datasets. 
Finally, we analyze the distribution of dataset attributes.

\noindent\textbf{Learning setting and notations.} We work in a supervised learning setting commonly used to build deep face representations~\cite{cao2018vggface2,kim2022adaface,kim2023dcface}.
Training datasets are defined as $\mathcal{D} = \{P_1, P_2,..., P_n\}$, with $P_k$ being an identity, real or generated. 
Each identity is represented as $P_k = (I_k, F_k)$, with $I_k$ a set of $n$ representative face images and $F_k = \{f_k^1, f_k^2,...,f_k^n\}$ a set of identity attributes.
These attributes can be demographic or visual characteristics of faces, as discussed below.

\subsection{Promoting Fairness in Face Recognition Training Datasets}
\label{subsec_balancing}
\begin{figure}
    \centering
    \includegraphics[width=\linewidth]{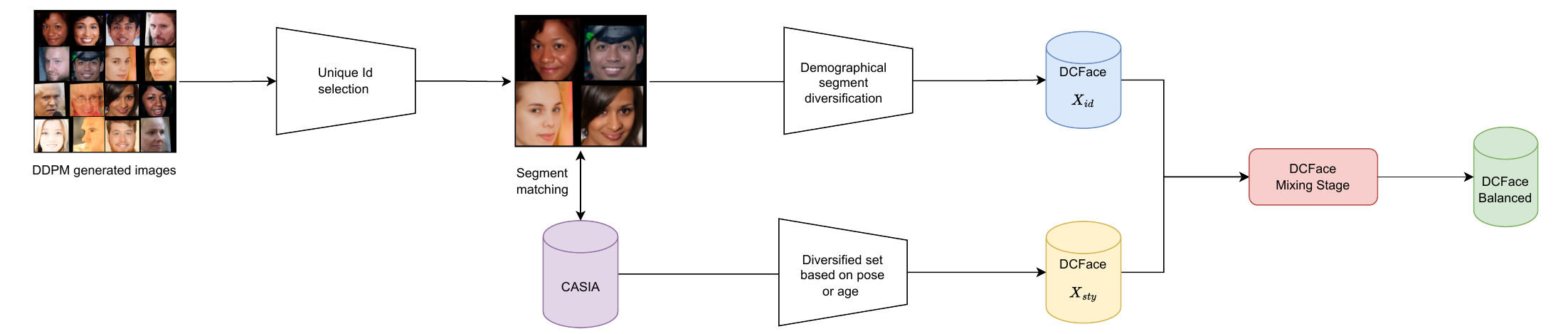}
    \caption{Proposed attribute control method applied to DCFace~\cite{kim2023dcface} incorporating attribute balancing. Ethnicity and gender are controlled when generating the ID image. Age and pose are controlled when mixing images to obtain the visual representation of each identity. The figure is a modification of the original one from~\cite{kim2023dcface}.} 
    \label{fig_control}
\end{figure}

We implement dataset balancing with different combinations of the following attributes used for balancing: \origin - ethnicity, \gender - gender, \age - age, and \pose - pose. 
They were selected due to their importance for demographic fairness for the first three and to face appearance variability for pose. 
\origin and \gender are attributes uniquely associated with each identity. 
\age and \pose are estimated for individual images and then aggregated.  
The proposed fairness promotion method is generic and can be applied to any dataset generation process.
In Figure~\ref{fig_control}, we exemplify the method with the modified version of DCFace~\cite{kim2023dcface} that enables attribute control to balance the training dataset.

\textbf{Demographic Balancing.}
The proposed method relies on ethnicity, gender, and age predictions obtained with FairFace~\cite{karkkainen2019fairface}. 
The following ethnic categories are used when balancing the datasets: Asian, Black, Indian, and White.
Gender predictions are binary (female/male), while age predictions are split into three categories: young (under 30 years old), adult (30-50 years old), and senior (over 50 years old). 
Ethnicity and gender are controlled when generating the seed image for each identity.
We aim to get close to a balanced representation of each demographic segment determined by $(\genderNosp,\originNosp)$ combinations.
The age varies for each identity and should be controlled at the individual image level. 
It is controlled for the style images that are mixed with the diffusion image seed to obtain the full representation of each identity. 

The used attribute representations are imperfect because there is no universally agreed way to define them. 
The discretization choices made are determined by the availability of resources needed to define or predict attributes.
They are informed by ethical guidelines regarding ethnicity~\cite{lee2008ethics} and gender~\cite{quinan2022biometric}.
The proposed approach is flexible and can be modified upon the future availability of better attribute representations.

\textbf{Pose balancing.}
Pose diversification and balancing builds on the pose estimation algorithm introduced in~\cite{hempel20226d}. 
This method leverages a 6D rotation matrix formalism to infer a robust pose estimation in a continuous representation space. 
From the pitch, the yaw, and the roll axes returned by the method (i.e. the rotations around respectively the $x$, $y$, and $z$ axes), we perform a classification of the pose to guarantee its diversity within the style images. 
This 3D classification is performed by creating N classes $C_{k, k\in [1,N]}$ for each rotation axis.
To maintain a minimum representativity for each class, we impose $N=3$. 
The class boundaries are then set with a dynamic thresholding that imposes $C_1$ and $C_3$ to be equally represented and the neutral class $C_1$ to contain $68\%$ of the images. 
This approach is due to the diverse shapes the pose distributions take (laplacian/gaussian-like) for the different axes, which is not favorable for thresholding based on close formula(function of standard deviation, equal-sized splits, etc). 
We finally define the final position classes as being the cartesian product of the previously computed classes  $P_{i,j,k} = (C_i^{x_{axis}}, C_j^{y_{axis}},C_k^{z_{axis}})$. The style images are thus diversified by evenly sampling the 27 resulting classes $P_{i,j,k}$. 
The obtained clusters are illustrated in the supplementary material. 
Given an initial set of images, we can enable pose balancing by assigning them to pose clusters and targeting a balanced distribution among clusters.
The pose should vary for individual samples and is controlled on the lower branch of the modified DCFace pipeline presented in~\autoref{fig_control}.

\subsection{Compared Training Datasets}
\label{subsec_frt-datasets}

We target a thorough comparison of face recognition training datasets.
To this end, we select or create real and generated resources that exemplify different ways to build training datasets. 
All datasets have a similar structure to ensure fairness of the comparison.
They include $10,000$ identities, with 50 representative images per identity, except for CASIA which has 49.4 images per identity on average.  
The following existing datasets cover the main recent tendencies in face generation, as well as the use of real images:
\begin{itemize}
\item CASIA~\cite{yi2014learning} - real dataset representing celebrities from the IMDB dataset. 
\item SynFace~\cite{qiu2021synface} - synthetic dataset created with a GAN architecture using identity and domain mixup to diversify generated faces. 
\item DigiFace~\cite{bae2023digiface} - synthetic dataset created using rendering technique to obtain diversified representations of faces of each identity.
\item \dcface~\cite{kim2023dcface} - a synthetic dataset generated using the pipeline of \cite{kim2023dcface} that uses a diffusion model to generate a diversified dataset. A seed image is generated to create a fictitious identity. The full representation of the identity is obtained by combining the seed image and CASIA images. Diversified CASIA images are used to push for diversity in the representation.  
\end{itemize}

We introduce the variants of DCFace that include control of one or several attributes following the procedure described in Subsection~\autoref{subsec_balancing}:
\begin{itemize}
    \item \dcfaceOG - a version of \dcface with gender and ethnicity balancing of the seed images of the fictitious identities. Gender and ethnicity are inferred with  FairFace~\cite{karkkainen2019fairface}.
    \item \dcfaceOGA - adds age control to \dcfaceOG by selecting images to make age distribution as even as possible across age ranges.
    \item \dcfaceAll - adds age and pose control to \dcfaceOG \dcfaceOG by selecting images to make joint age-pose distribution as even as possible.
\end{itemize}

\begin{table}
\definecolor{color0}{RGB}{247.22348176337726,200.63106573360696,208.80330014218268}
\definecolor{color1}{RGB}{189.1482947660368,239.05375800818038,166.56063484759653}
\definecolor{color2}{RGB}{244.52347303953013,222.59477567045175,226.40064691644085}
\definecolor{color3}{RGB}{255,255,255}
\definecolor{color4}{RGB}{173.99760206330618,237.94103887815803,144.66740217950152}
\definecolor{color5}{RGB}{243.95304866125255,227.2349960796443,230.11839623494612}
\definecolor{color6}{RGB}{244.27628914227654,224.60553784776852,228.01167162112645}
\definecolor{color7}{RGB}{139.52977616459395,235.40960285735716,94.86029785958534}
\definecolor{color8}{RGB}{242.9833272181807,235.12337077527167,236.43857007640506}
\definecolor{color9}{RGB}{170.96746352276003,237.71849505215354,140.2887556458825}
\definecolor{color10}{RGB}{245.22699643940578,216.87183716578093,221.81542275695105}
\definecolor{color11}{RGB}{244.3143174341617,224.29618982048902,227.76382166655947}
\definecolor{color12}{RGB}{244.4474164557598,223.21347172501075,226.8963468255749}
\definecolor{color13}{RGB}{242.25,242.25,242.25}
\definecolor{color14}{RGB}{242.4509311317883,239.45424315718472,239.90846944034328}
\definecolor{color15}{RGB}{246.19671788247766,208.98346247015357,215.4952489154921}
\definecolor{color16}{RGB}{244.86572766649667,219.81064342493622,224.1699973253377}
\resizebox{0.99\linewidth}{!}
{
\begin{tabular}[tb]{cccccccc}
\hline
method & \casia & \digiface & SynFace  & \dcface & \textbf{\dcfaceOGNosp} & \textbf{\dcfaceOGANosp} & \textbf{\dcfaceAllNosp}  \\
\hline
\cellcolor{color3} \textbf{g} & \cellcolor{color7} 99.70 & \cellcolor{color7} 93.04 & \cellcolor{color7} 99.24 & \cellcolor{color7} 98.87 & \cellcolor{color7} 99.99, ff & \cellcolor{color7} 99.94, ff & \cellcolor{color7} 99.80, ff \\
\cellcolor{color3} \textbf{e} & \cellcolor{color0} 46.90 & \cellcolor{color14} 65.32 & \cellcolor{color0} 39.67 & \cellcolor{color15} 55.57 & \cellcolor{color9} 92.82, ff & \cellcolor{color4} 92.22, ff & \cellcolor{color1} 90.00, ff \\
\cellcolor{color3} \textbf{a} & \cellcolor{color16} 58.72 & \cellcolor{color0} 42.45 & \cellcolor{color8} 63.66 & \cellcolor{color11} 64.11 & \cellcolor{color10} 60.66 & \cellcolor{color2} 65.00, ff & \cellcolor{color6} 69.39, ff \\
\cellcolor{color3} \textbf{p} & \cellcolor{color12} 60.85 & \cellcolor{color13} 67.03 & \cellcolor{color5} 58.35 & \cellcolor{color0} 50.84 & \cellcolor{color0} 51.39 & \cellcolor{color0} 52.93 & \cellcolor{color0} 57.51, cl \\
\hline
\end{tabular}
}

\caption{ Degree of balance for the compared datasets. \fface indicates that FairFace is used for balancing. \textit{cl} indicates that clustering-based pose diversification was applied. The degree of balance is quantified by the entropy of the representativity of each class for a given attribute. \textbf{Datasets in bold are introduced here}.}
\label{tab_datasets}

\end{table}

\subsection{Face Attributes Analysis}
\label{subsec_analysis}

Real dataset creation processes~\cite{cao2018vggface2,guo2016ms,kim2023dcface} focus on obtaining a representative set $I_k$.
Surprisingly, little attention is paid to demographic and visual attributes balancing. 
The resulting training datasets collected from the Web are usually unevenly distributed.
We highlight these biases in Table~\ref{tab_datasets} by color-coding the degree of imbalance using FairFace predictions of the attributes. 
We note that ethnicity and gender are globally easier to control than age and pose. 
This difference is explained by the fact that diversification is applied at a coarser granularity level for the first two attributes. 
Age and pose are distributed using a limited amount of available photos per identity. 
 
Face image generation is, in principle, easier to control than Web-based collection.
However, the results presented in Table~\ref{tab_datasets} show that there are important biases in existing datasets. 
\digiface implements a rendering pipeline and ensures a better balancing than \dcface for ethnicity and pose but is less equitable for gender and age. 
\dcfaceAll ensures the best distribution of demographic attributes among generated datasets, indicating that the proposed balancing method is effective.

\section{Experiments}
\label{sec_experiments}
We compare the different training datasets in a face verification task using seven datasets.
After briefly presenting the FRT training pipeline, we introduce the evaluation datasets and their enrichment with attribute-related metadata to enable a systematic evaluation of fairness in addition to accuracy.
We then discuss the obtained results, including a statistical analysis that reveals the role of identity attributes. 

\noindent\textbf{Implementation}
All models are trained with the AdaFace method~\cite{kim2022adaface} to facilitate comparability of results.
The backbone network is a ResNet50~\cite{he2016deep} and hyperparameters are the same as~\cite{kim2022adaface}.
A detailed list of parameters is provided in the supplementary material. 

\subsection{Evaluation Datasets, Protocol and Metrics}
\label{subsec_eval-datasets}
We evaluate the deep representation obtained with the compared datasets in the widely used face verification task~\cite{taskiran2020face}.
It aims to assert whether two images of faces represent the same person. 
Recent works~\cite{bae2023digiface,kim2022adaface,kim2023dcface,qiu2021synface} evaluate performance on various datasets, which the most popular are:  (1) Labeled Faces in the Wild (\lfwNosp)~\cite{LFWTechUpdate}, the reference dataset for the task (2) \calfwNosp~\cite{zheng2017calfw}, a version of \lfwNosp with a larger age variability, (3) \cplfwNosp~\cite{zheng2018cplfw}, a version of \lfwNosp with pose variability, (4) \agedbNosp~\cite{moschoglou2017agedb}, a dataset designed for maximizing age variability, and (5) \cfpfpNosp~\cite{sengupta2016cfp} that is designed for pose variability, (6) \lfwc - a more challenging version of \lfw with a selection of negative pairs using the nearest identity from the dataset computed using the pre-trained IR50 model~\cite{zhao2018towards}, and (7) \favcidNosp~\cite{popescu2022face} a challenging dataset that is more balanced and larger than \lfw, with \origin (country of origin) and \gender diversification based on Wikipedia metadata. 
We do not include RFW~\cite{wang2019racial} because this dataset is sourced from MS1M~\cite{guo2016ms}, a dataset that was withdrawn and should not be used.


We follow the standard evaluation protocol~\cite{LFWTechUpdate}.
We use the accuracy obtained when applying this procedure for each verification dataset to evaluate each compared dataset and provide a more detailed analysis in~\autoref{sec_analysis}.

\begin{table}[bt]
\begin{center}
\resizebox{0.99\linewidth}{!}
{
\begin{tabular}{|c|c|c|c|c|c|c|c|}
\hline
\multirow{2}{*}{\makecell{Verif.\\ dataset}} & \multicolumn{1}{c|}{Real dataset} & \multicolumn{6}{|c|}{Synthetic datasets} \\ 
\cline{2-8}
& \casia & SynFace & \digiface & \dcface & \dcfaceOGNosp & \dcfaceOGANosp & \dcfaceAllNosp \\
\hline
\lfwNosp & 99.46 & 87.28 & 89.08 & 98.13 & 98.19 & 98.0 & \textbf{98.5}\\
\hline
\cfpfpNosp & 94.87 & 67.01 & 78.45 & 80.92 & 80.58 & 80.77 & \textbf{81.57}\\
\hline
\cplfwNosp & 90.35 & 64.91 & 70.28 & 79.94 & 79.56 & 79.43 & \textbf{80.31}\\
\hline
\agedbNosp & 94.95 & 61.78 & 65.71 & \textbf{87.96} & 87.26 & 87.08 & 86.48\\
\hline
\calfwNosp & 93.78 & 73.53 & 71.7 & 90.39 & 89.91 & 90.66 & \textbf{90.68}\\
\hline
\lfwcNosp & 89.39 & 61.74 & 56.16 &\textbf{77.67}  &  73.01 &  73.7& 77.4   \\
 \hline
\favcidNosp &  81.7  & 63.64 & 61.19 & 72.84 & 73.53  & 73.51 & \textbf{74.09}   \\
\specialrule{.2em}{.1em}{.1em}
AVG & 92.07 & 68.56 & 70.37 & 83.98 & 83.14 & 83.31 &  \textbf{84.15} \\
 \hline
 
\end{tabular}
}
\end{center}
\caption{Accuracy for seven face verification datasets and deep representations trained with the face recognition datasets introduced in~\autoref{subsec_frt-datasets}. Higher values are better. \textbf{Best synthetic results in bold} (those on the real dataset are always the overall best).}
\label{tab_error}
\end{table}

\subsection{Face Verification Accuracy}
\label{subsec_accuracy}
~\autoref{tab_error} presents the face verification accuracies for models obtained with training datasets described in~\autoref{subsec_frt-datasets}. 
Compared to previous works \cite{qiu2021synface, kim2023dcface,bae2023digiface}, we add two more verification sets \favcid \cite{popescu2022face} and \lfwc{}, the only difference between the latter and \lfwcNosp{} being a more challenging image pairs rearrangement.
When testing on these two sets, the results are less optimistic regarding the performance of synthetic datasets than the conclusion of~\cite{kim2023dcface}. The obtained accuracies show that an important gap in favor of models trained with real images still exists.  
Moreover, they show that the accuracies on current verification sets are saturating and are thus less suitable for performance comparison. This issue could be tackled in future works by evaluating on more challenging image pairs similar to \lfwcNosp.

However, the results obtained for synthetic datasets still confirm previous findings regarding the progress made using very recent training datasets, such as DCFace~\cite{kim2023dcface}. 
The proposed \dcfaceAllNosp{} brings further improvements and obtains the best result. Drops in accuracy for \dcfaceOGNosp{} and \dcfaceOGANosp{} seem to be the consequences of single attributes control on the resulting dataset pose or age diversity as shown in  \autoref{tab_datasets}.

This finding supports the usefulness of the proposed diversification, which improves accuracy and fairness, further analyzed in details in the next section.


\section{Analysis}
\label{sec_analysis}
\subsection{Motivation and description}
 
Let us consider $n$ sampled data points $(x_i,y_i)\overset{i.i.d.}{\sim}(X,Y)$, and predictions $\hat{y_i} = f_\theta(x_i)$ with the predictor $f_\theta$. We aim to maximize the expectation of a metric $m$ (error rate, accuracy, specificity, AUC, ...), $\mathbb{E}[m(Y, \hat{Y})]$. We say that $f_\theta$ is fair on the metric $m$ between protected population attributes $s$ and $s'$ if 
\begin{equation}
\label{eq_equal_op}
    \mathbb{E}[m(Y, \hat{Y}) | s] = \mathbb{E}[m(Y, \hat{Y}) | s']
\end{equation}

For a binary metric $m$ (for example True Positive Rate TPR, $m(y, \hat{y}) = \mathds{1}[y=\hat{y}|y=1]$), we have $\mathbb{E}[m(Y, \hat{Y}) | s] = \mathbb{P}[m(Y, \hat{Y}) | s]$. Therefore,  \autoref{eq_equal_op} is a generalization of \textit{equal opportunity} as defined in \cite{hardt2016equality_opportunity} that can be adapted to any binary metric.

\begin{figure}[t]
    \centering
    \includegraphics[width=\linewidth]{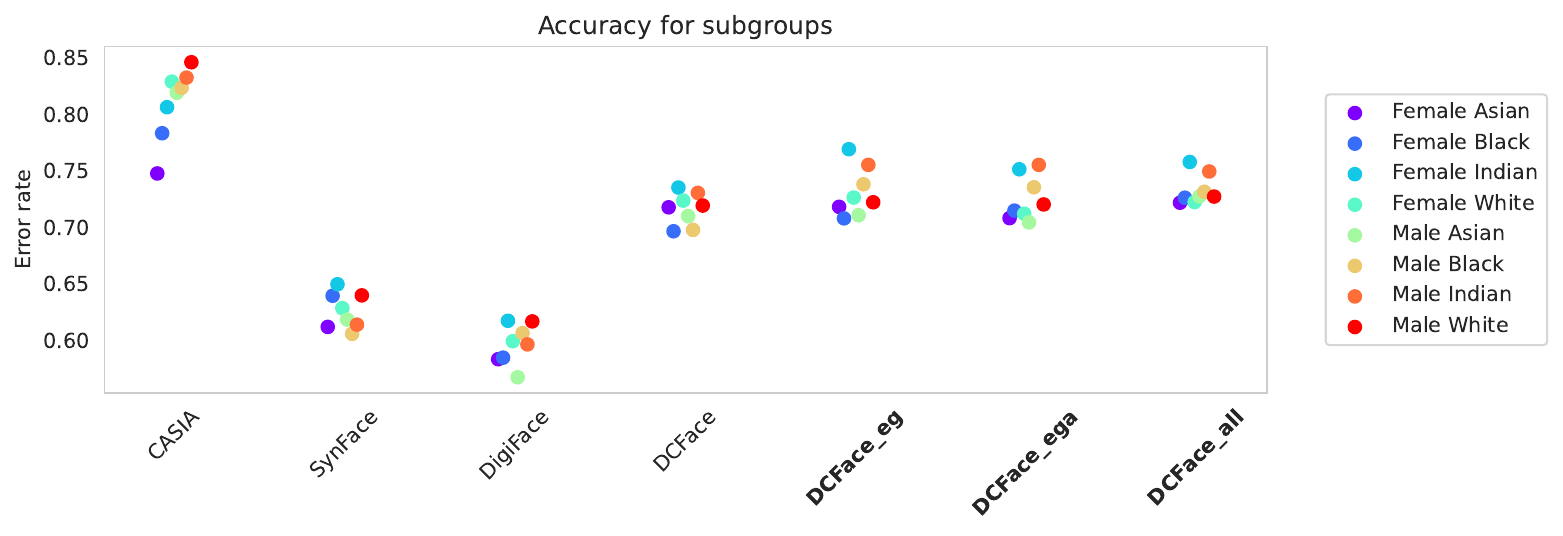}
    \caption{Accuracy imbalance for \favcid}
    \label{fig_scatter}
\end{figure}

However, one can only approximate $\mathbb{E}[m(Y, \hat{Y}) | s]$ and $\mathbb{E}[m(Y, \hat{Y}) | s']$ by their empirical counterpart. For the accuracy metric, the disparities between subgroups of \favcid are presented in \autoref{fig_scatter}. Without further analysis, it is impossible to conclude on the fairness of a model, due to two major points:
\begin{enumerate}
    \item Random effects: for under-represented subgroups, the estimate of the expectancy can be quite far from its theoretical counterpart. Moreover, there may be a different intrinsic variance in each subgroup.
    \item Co-founding factors: if a subgroup is consistently associated with another variable (such as a bigger pose variation), then we need to control for it, in order to isolate the true marginal effects.
\end{enumerate}

Therefore, a finer analysis is needed to properly measure for which subgroups $s$ and $s'$ we have $\mathbb{E}[m(Y, \hat{Y}) | s] \neq \mathbb{E}[m(Y, \hat{Y}) | s']$. Using the \textit{accuracy} for the metric $m$ would lead to a limited analysis, thus we also consider TPR (True Positive Rate) and FPR (False Positive Rate) metrics, as proposed in \cite{sarridis2023fair}. Hence, we will separately measure unfairness on the positive and negative image pairs. 

We also want to identify the primary factors influencing latent representations and measure performance discrepancies attributed to \textit{protected attributes} (such as \textit{gender}, \textit{age} and \textit{ethnicity}). Following common statistical and econometrical practices \cite{angrist2009mostly, gareth2013introduction}, we employ LOGIT (logistic) regressions and ANOVA (ANalysis Of VAriance) to identify statistically significant results and interpret the predictive margins and uncertainty quantities. 
More precisely, for each image pair, we consider two target variables: \textit{dist} $\in [0, 2]$, which is the Euclidean distance between the normalized latent representations of the images composing the pair; and the binary prediction $\mathbf{\hat{y}}$, which is equal to 1 if the model predicts the same person on both images, and 0 otherwise. We denote by $\mathbf{y}$ the ground truth variable.

The explanatory categorical variables are the gender combination in the pair $\mathbf{g} = \mathbf{g_1 \times g_2}$, the age combination $\mathbf{a} = \mathbf{a_1 \times a_2}$, and the ethnicity combination $\mathbf{e} = \mathbf{e_1 \times e_2}$. 
These combinations are not ordered in a pair, such that \eg $\{Male \times Female\} =\{Female \times Male\}$. 
Additionally, we have $\mathbf{p}^{x}_{1,2}$, $\mathbf{p}^{y}_{1,2}$, $\mathbf{p}^{z}_{1,2}$ the head pose 3D angles between the two faces. 
We also have an additional variable $D$ to model training set used to generate \textit{dist} and $\mathbf{\hat{y}}$.

We perform multiple statistical regressions, detailed in the following subsections, jointly on all our variables. This allows us to correct the effect of possibly correlated variables. For example, if the users tagged as ``\textit{White}'' are also more frequently tagged as ``\textit{Senior}'', regressions that only include one or the other will not lead to interpretable causal effects. Regressions on all explanatory variables help us to isolate each marginal effects.

\subsection{Marginal Variable Analysis}
\label{subsec_variable}
\begin{figure}[t!]
    \centering
    \includegraphics[width=\linewidth]{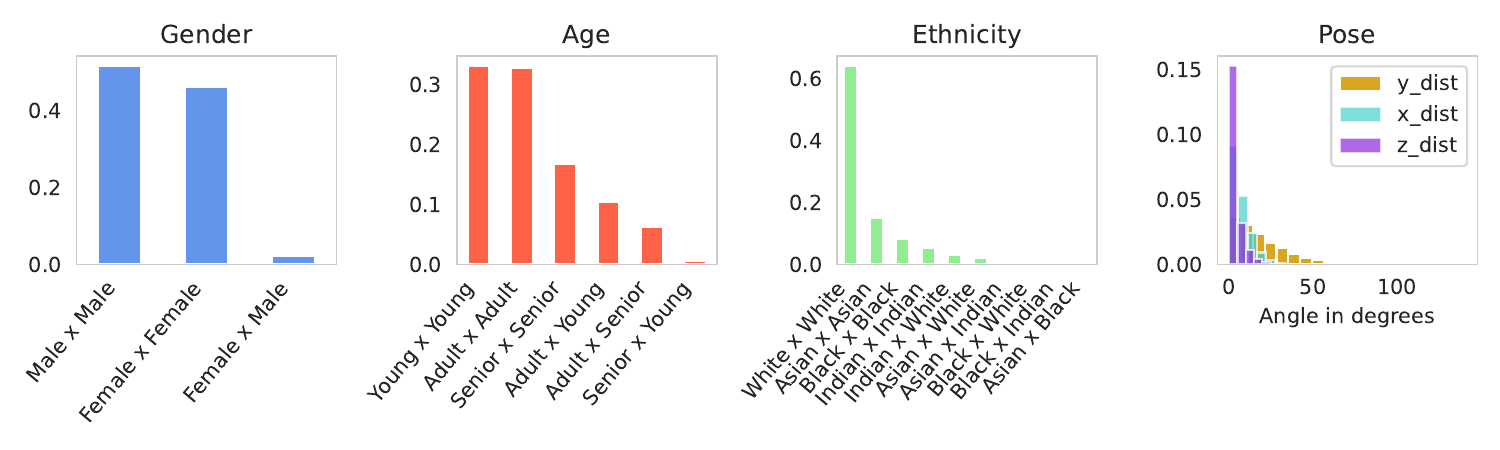}
    \caption{Distribution of the pair attributes in \favcid}
    \label{fig_variables}
\end{figure}

To perform a complete analysis of the models fairness, we will evaluate the fairness with the FAVCI2D dataset. Since this dataset provides a high number of balanced pairs of image to evaluate, it is the most suitable to analyze the performances on subgroups, since finer analyses required a sufficient number of samples for each class.  
First, we shall look at the dataset's protected attribute combination distribution in \favcid. As shown in \autoref{fig_variables}, most image pairs represent the same gender. However, there is a higher class imbalance for the age pairs and an even higher one for ethnicity pairs. Moreover, most pairs contain the same attributes in both images, creating a more challenging setup. It is to be noted that when different ethnicities appear in a pair, one of the images often represents a white person. Therefore, in \favcidNosp, it is difficult to accurately measure biases regarding two distinct non-white ethnic groups. We will denote by ``soft'' pairs the pairs composed of images with different attributes and ``hard'' pairs the pairs composed of images with identical attributes.  Spatial variability comes from the $y$-axis, and the most lack of variation comes from the $z$-axis.

In the following, we will focus our analysis on the ``hard'' pairs since they are the ones most likely to exhibit unfairness in the models.

\subsection{Representation Space Analysis}
\label{subsec_representation}

\begin{figure}[b]
    \centering
    \includegraphics[width=\linewidth]{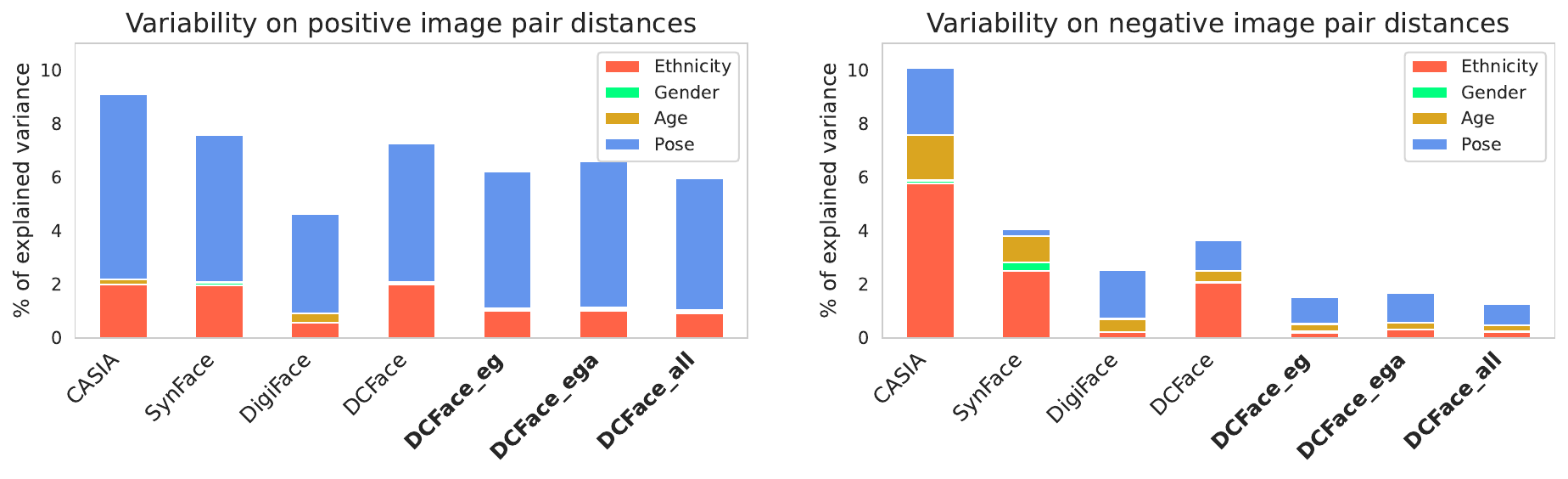}
\caption{ANOVA results: total height corresponds to $R^2$, the explained variance by the variables. Each bar is decomposed into multiple $\eta^2$,  the individual contributions to the variance}
\label{fig_ANOVA}
\end{figure}

We then perform a linear regression of the distance between image pair embeddings and our variables of interest. The coefficient of determination $R^2$ of the regression can be interpreted as the variance of the distances explained by the variables, and a further ANOVA \cite{gareth2013introduction} is used to obtain aggregated statistics on the categorical and continuous variables. In particular, we look at the individual contributions of the variance $\eta^2$ derived from the ANOVA. To interpret these quantities as ``explained variance'', we first check the normality and homoscedasticity\footnote{Constant variance} of the residuals\footnote{The residuals are the difference between predicted and real target variable} of the regression. These assumptions are achieved by converting the Euclidean distance between image pair embeddings to an angle in degrees. For simplicity, we consider the global angle between the head poses $\mathbf{p_1}$ and $\mathbf{p_2}$ instead of the contribution of each individual axis.
The results can be found in~\autoref{fig_ANOVA}.

For positive pairs, the models' outputs are significantly influenced by the difference in pose between the two images. This is coherent with the data in~\autoref{tab_datasets}, and would likely illustrate limited pose variation within the datasets, especially along the $z$ axis (\autoref{fig_variables}). Although ethnicity does have some impact, it is relatively minor when compared to the variance due to pose differences. This suggests that the models are adept at mapping images of individuals with the same ethnicity — and potentially the same gender and age — to similar sub-regions within their latent space. Such patterns are visually confirmed through the use of dimensionality reduction techniques like UMAP, with corresponding images to be found in the supplementary material. 

A greater fraction of variance can be attributed to ethnicity and age for the negative pairs, with very different behaviors between different training datasets. Unbalanced datasets tend to populate the embedding space unequaly, creating a dependence between the embedding's distance and protected attributes. The standard \dcface, \casia, and SynFace models display susceptibility to this effect. Conversely, our modified \dcface versions show a significantly decreased sensitivity to these attributes, indicating progress toward more equitable model behavior.

Overall, conditioning the generation on protected attributes seems to enhance the results on \favcid mainly because the negative class distances become much less dependent on such attributes.


\subsection{TPR and FPR analysis}
Looking at the distances in the latent space is a good indicator of the information that each model uses to project each image. 
However, the literature discusses accuracy, recall, and precision metrics because they are much more interpretable. 
Therefore, for each image pair, we create a variable $\mathbf{\hat{y}} = \mathds{1}\{\textit{dist} \leq \textit{thresh}\}$ where \textit{thresh} is the threshold selected to maximize the global accuracy of the evaluation dataset. The individual target variable becomes $\mathds{1}\{ \mathbf{y} = \mathbf{\hat{y}}\}$, which is a binary variable modeling a correct identification of an individual. Our goal is to estimate the performances of each subgroup: $\mathbb{P}[  \mathbf{y} = \mathbf{\hat{y}} | \mathbf{e}=e]$, and compare them for different values of $e$, all while controlling for other variables. As in the previous section, we distinguish between positive and negative pairs, which is equivalent to modeling the TPR and FPR, respectively.

\subsubsection{Model.}
Since our target variable is binary, we perform a LOGIT regression \cite{angrist2009mostly}. Let us denote by $X$ all of our binary and continuous explanatory variables. The Logit model fits a regression of the form:
\begin{equation*}
  \ln \dfrac{\mathbb{P}[y=1|X]}{\mathbb{P}[y=0|X]} = \beta_0 + \beta_1 X_1 + ... +\beta_k X_k 
\end{equation*}
where $y$ is a target binary variable to explain.
In other terms, it models the log-odd-ratio of the outcome conditionally to the predictors. Once fitted, the estimated $\hat{\beta_i}$ can be used to estimate the probability $\hat{\mathbb{P}}[y=1| \Tilde{X}]$, where $\Tilde{X}$ can be a possibly unknown combination of predictors. The coefficients $\hat{\beta_i}$ are estimated with a confidence interval, and one should check if they are significantly different from 0. Categorical variables are encoded using dummy coding: we fix a reference category level (for example the performance of ``\textit{White}`` for the category $\textbf{e}$), and the coefficients of the other levels represent variations \textit{with respect} to the reference level. In particular, we can calculate the marginal effect of each level compared to the reference level. If the marginal effect is positive, with a $p$-value over .05, it means that the associated level is significatively associated with an increase in $y$ compared to the reference level. This methodology allows us to model and measure significative differences between different subgroups of data.

\subsubsection{Results.}
\begin{figure}[t]
    \centering
    \includegraphics[width=\linewidth]{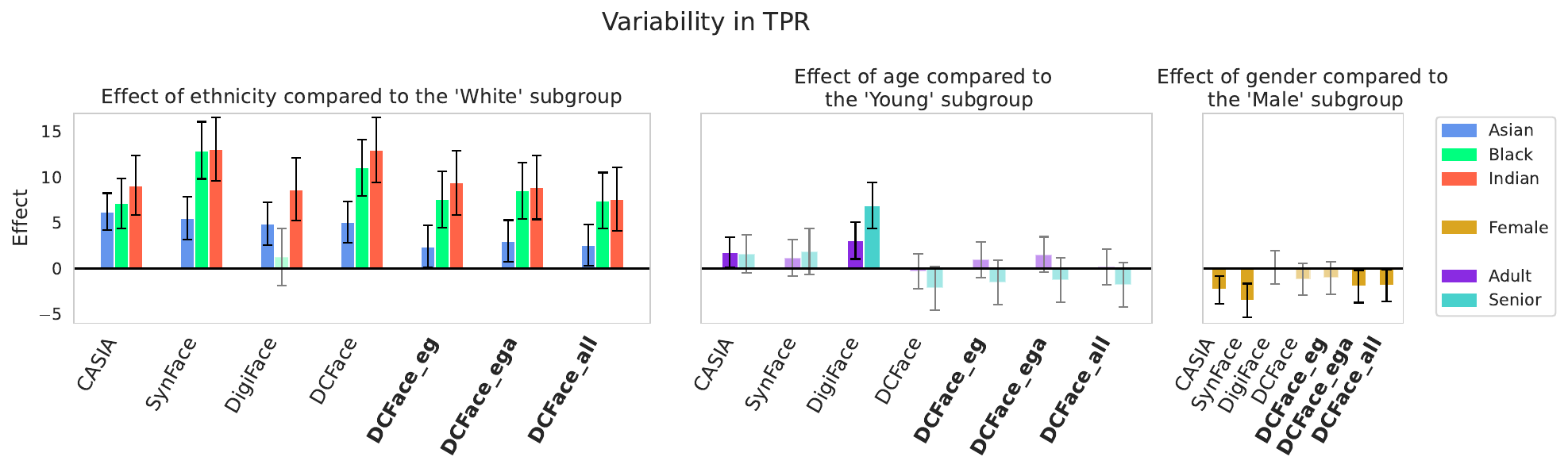}
    \caption{Marginal effect on TPR of each variable combination for each method on the ``hard'' pairs versus the reference combination.} 
    \label{fig_TPR}
\end{figure}

\begin{figure}[t]
    \centering
    \includegraphics[width=\linewidth]{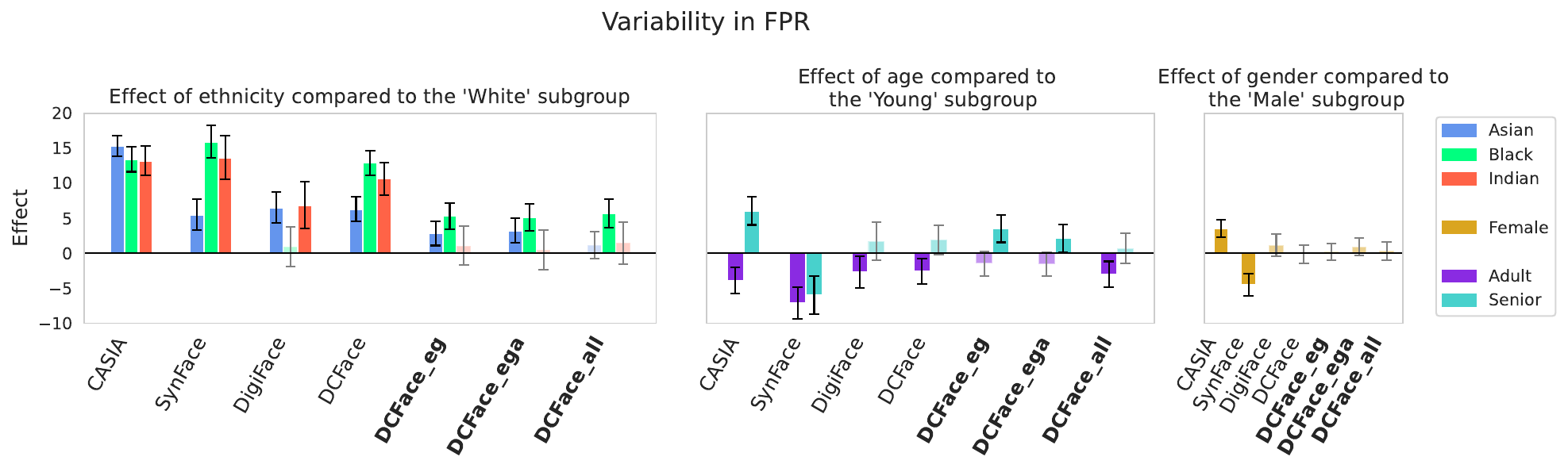}
    \caption{Marginal effect on FPR of each variable combination for each method on the ``hard'' pairs versus the reference combination.} 
    \label{fig_FPR}
  
\end{figure}

We chose their most present occurrence in the dataset for each categorical pair variable as a reference level. Therefore, the reference gender pair will be $\textit{Male} \times \textit{Male}$, the reference age pair will be $\textit{Young} \times \textit{Young}$, and the reference ethnicity pair will be $\textit{White} \times \textit{White}$.
We perform one logistic regression for each method and look for the marginal effects of each level of a category with respect to their reference level, as presented in~\autoref{fig_TPR} and \ref{fig_FPR}. Only non-transparent bars are significant at the .05 confidence level. 
As we see in~\autoref{fig_TPR}, with respect to ethnicity, all models exhibit some kind of unfairness regarding the TPR metric. In particular, all subgroups correlate with a better identification than the ``\textit{White}`` subgroup. This could be due to the poor coverability of the latent space by these subgroups, resulting in a high recall. Nevertheless, we see that modified versions of \dcface do seem to have reduced TPR dependence on ethnicity. With respect to the FPR metric (Figure \ref{fig_FPR}), however, the results are much more interesting: our modified versions of \dcface show greater fairness between ethnic groups. In particular, \dcfaceAll shows fairness between the ``\textit{Asian}`` and  ``\textit{White}`` subgroups, and between the ``\textit{Indian}`` and  ``\textit{White}`` subgroups, all while 
 greatly reducing the unfairness between the ``\textit{Indian}`` and  ``\textit{Black}`` subgroups. This is done while not reducing the fairness of other protected attributes. Even if our modified \dcface methods still exhibit some degree of unfairness, it seems that \dcfaceOGA has less age bias than \dcfaceOG, which shows the importance of controlling multiple attributes at once.

As we see, controlling the generation of synthetic images does lead to more fairness on multiple metrics, compared to synthetic and non-synthetic datasets.

\section{Conclusion}
\label{sec_conclusion}
The proposed work contributes to mitigate biases in face recognition and verification. 
The attribute-balancing approach applied during training data design improves fairness downstream during face verification. 
We systematically consider fairness in face verification evaluation and introduce a rigorous analysis framework to highlight the role of individual attributes and their combinations. 
Globally, we advocate for a transparent development of face recognition and verification.
This is necessary to advance research in the field and enable an open auditing of the commercially deployed systems. 
Such auditing improves transparency and can increase the social acceptability of these technologies. 
We show that a finer-grained representation is desirable for gender, ethnicity, and age, but it requires the creation of dedicated resources.

Second, while generation approaches reduce the role of real data, the latter are still needed, in particular for more challenging dataset than LFW. 
\digiface relies the least on actual faces, but its accuracy is clearly behind that of \dcfaceNosp.
This last approach needs CASIA images to obtain the final fictitious identity representations. 
In the future, we will look at ways of implementing the 'style' branch of the method with generated images.
Beyond privacy, this will further increase the control of the style images, and a finer-grained control of the entire dataset generation process. 

Third, the deep representations learned are still based on a supervised learning approach. 
The structure of the existing training datasets partly determined this modeling choice to ensure comparability.
The obtained results are interesting, but performance can be further improved by exploiting larger datasets that are structured differently. 

\section*{Acknowledgement}
The following publications report results funded in part by France 2030 program, managed by the ANR, project ANR-23-PEIA-0008, in the context of the PEPR IA.
This publication was provided with computer and storage resources by GENCI at IDRIS thanks to the grant 2023-AD011014868 on the supercomputer
Jean Zay's the CSL, V100 and A100 partition. 

\bibliographystyle{splncs04}
\bibliography{egbib}

\newpage
\title{Supplementary material to: Toward Fairer Face Recognition Datasets}
\titlerunning{Fairer Face Datasets}

\author{Alexandre Fournier-Mongieux$^*$\inst{1}\orcidlink{0009-0002-7744-3179} \and
Michael Soumm$^*$\inst{1}\orcidlink{0009-0009-0435-9903} \and
Adrian Popescu\inst{1}\orcidlink{0000-0002-8099-824X} \and
Bertrand Luvison\inst{1}\orcidlink{0000-0003-2475-3712} \and
Hervé Le Borgne\inst{1}\orcidlink{0000-0003-0520-8436}
}

\authorrunning{A.~Fournier-Montgieux et al.}

\institute{Université Paris-Saclay, CEA, List, F-91120, Palaiseau, France
}

\maketitle
\appendix
\def\thefootnote{*}\footnotetext{These authors contributed equally to this work}\def\thefootnote{\arabic{footnote}}
\section{Parameters for training and generation}
For training the face classifier, we use the Adaface training pipeline \cite{kim2022adaface}. We apply the same augmentations, crop and low-resolution augmentations, for all training sets. We perform the training on 4 GPUs with a batch size of 256 (i.e. 64 per GPU),  the optimizer is the standard SGD with a learning rate of 0.1 and a momentum of 0.9. We use as a scheduler a multi-step learning rate decay whose milestones are the epochs 12,20,24 and the decay coefficient is 0.1. The training loss is that of Adaface~\cite{kim2022adaface}.  The margin parameter m is set to 0.4 and the control concentration constant h to 0.333 as recommended by \cite{kim2022adaface}. On each training set, the training lasts 60 epochs.

For generating the DCFACE set and its variants we use the generation pipeline of \cite{kim2023dcface}. 
We impose the $X_{id}$ image and the $X_{sty}$ to be of the same demographic group as we found that mismatching is likely to induce non-convergence of the resnet50 model when training on the resulting dataset (in particular when mismatching in gender). 
Randomly sampling the style image within the CASIA dataset thus results in a non-decreasing loss of the resnet network. Within the code of \cite{kim2023dcface}, there is a sampling strategy we haven't tested which consists of combining DDPM images with the closer CASIA faces. 
This approach was and still is unfortunately non-usable due to incomplete critical files \footnotemark{}. \footnotetext{ The provided center\_ir\_101\_adaface\_webface4m\_faces\_webface\_112x112.pth file doesn't have a required "similarity\_df" field.  Also, the dcface\_3x3.ckpt file doesn't seem to store the following property: recognition\_model.center.weight.data }
Moreover, this strategy is not mentioned in the original paper and, since it combines similar CASIA and DDPM faces in a resnet100 latent space, it seems to be in contradiction with what is stated within the ID Image Sampling subsection of \cite{kim2023dcface}.
We thus chose to ignore this strategy, our study being primarily an analysis of fairness and improvement research in this regard.

For all methods, similarly to what the original paper did, we introduce variability within the considered DDPM $X_{id}$ pictures by using a similar $F_{eval}$ model as in \cite{kim2023dcface}. However, one should be aware that the Cosine Similarity Threshold might vary depending on the training of the $F_{eval}$ network. We used the network trained on \cite{zhu2021webface260m} provided by the Adaface Github repository and found 0.6 as an effective threshold to filter too similar images. We also get rid of faces wearing glasses with the following solution \cite{Birskus_Glasses_Detector_2024}.

\section{LFW-C dataset}

We introduced in the main paper a variant of LFW, named LFW-C, which focuses on harder sample pairs. We explain here the pipeline to produce such an evaluation set. We first consider a resnet trained on a face classification task. We extract from this model the embeddings of all pictures within the original evaluation set pair file. We then create a new pair file with exactly the same number of positive and negative pairs as the original pair file such that:

\begin{itemize}
    \item The negative pairs are the list of different identities pairs of images composed by the closest pictures in terms of distance between embeddings.
    \item  The positive pairs are the list of same identity pairs of images composed by the most distant pictures in terms of distance between embeddings.
\end{itemize}

More formally and still considering only the images mentioned in the original pair file. 
Let $(NewNegPairs_i)_{i \in \mathbb{N}}$ be the new list of negative pair. We note by $F$ the embedding extraction function. The evaluation pairs list is built so that it respects the following assertion:

\begin{align}
    &\forall NegPair=(Pic_1, Pic_2) \not \in NewNegPairs, \\
    &\forall NewNegPair=(Pic'_1, Pic'_2) \in NewNegPairs, \\
    &||F(Pic'_1) - F(Pic'_2)|| \leq ||F(Pic_1) - F(Pic_2)||
\end{align}

Let $(NewPosPairs_i)_{i \in \mathbb{N}}$ be the new list of positive pairs. The list is built so that it respects the following assertion: 

\begin{align}
&\forall PosPair=(Pic_1, Pic_2) \not \in NewPosPairs,\\
&\forall NewPosPair=(Pic'_1, Pic'_2) \in NewPosPairs,\\
&||F(Pic'_1) - F(Pic'_2)|| \geq ||F(Pic_1) - F(Pic_2)||
\end{align}

One should be aware that it is crucial to consider only the pictures mentioned by the original pair file since they are more likely to not be subject to labeling mistakes. One has to be sure that the labels are correct before considering the entire evaluation set since this pipeline would then be, by construction, a magnet for labeling errors.



\section{Latent spaces analysis}
\begin{figure}[ht]
    \centering
    \includegraphics[width=\linewidth]{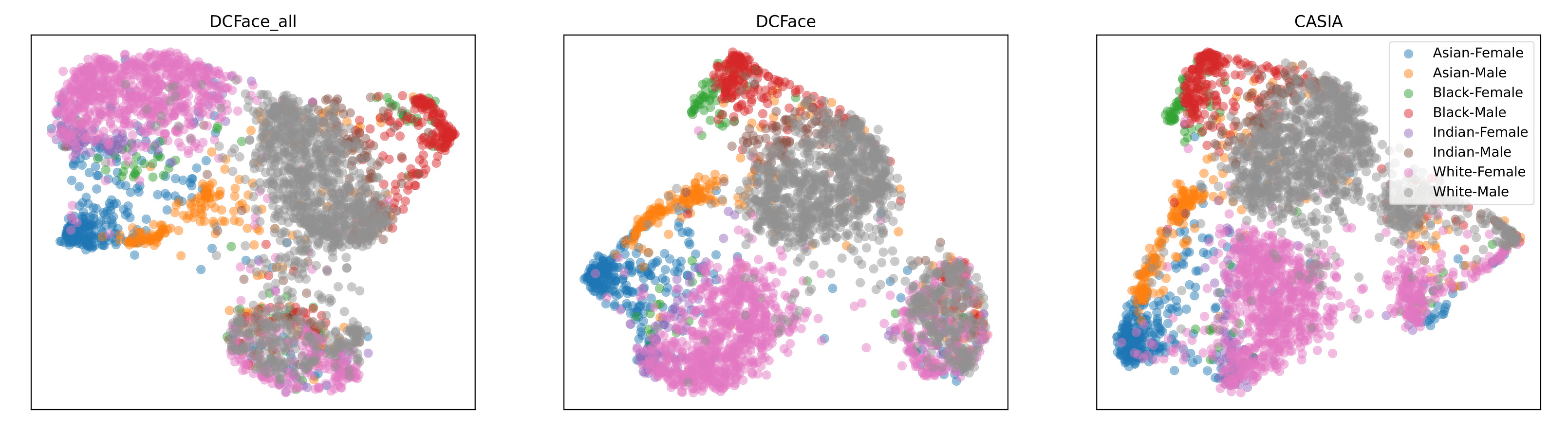}
    \caption{UMAPs \cite{mcinnes2020umap} applied on $DCFace$, $DCFace_{all}$ and CASIA }
    \label{fig_umaps}
\end{figure}

\begin{figure}[ht]
    \centering
    \includegraphics[width=\linewidth]{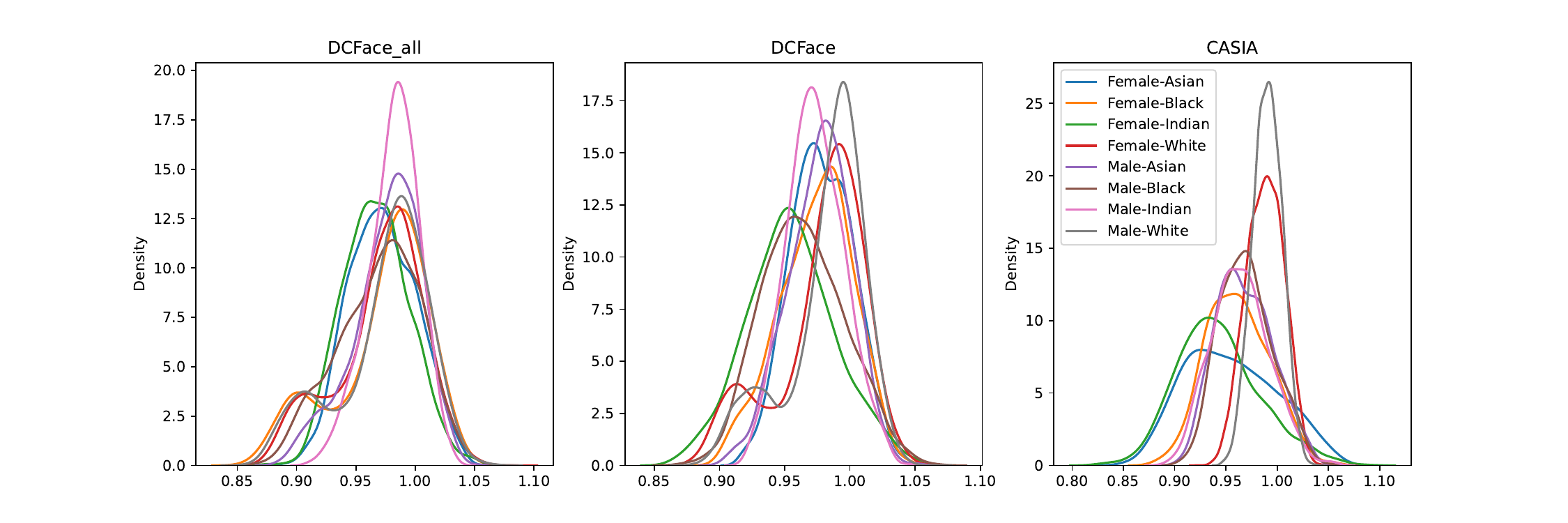}
    \caption{Distance distribution to the cluster mean for \favcid images. One image per identity is considered within each clustered}
    \label{fig_embeddings_dits}
\end{figure}

We show in \autoref{fig_umaps} the resulting application of UMAPs \cite{mcinnes2020umap} on $DCFace$, $DCFACE_{all}$ and $CASIA$. These figures are the first way to apprehend the characteristics of models to form clusters for the different ethnicities. We, however, observe that introducing controls in the synthetic dataset introduces less compact clusters than uncontrolled ones since the clusters are more scattered. While UMAP projection is locally pertinent to study structures in latent space, we also provide the distributions of the distances to the cluster mean for each subgroup in  \autoref{fig_embeddings_dits} as a clear visualization of the embeddings quality. We thus confirm that the data structure becomes more organized across the ethnicities by imposing control on the synthetic train set. While DCFace and CASIA show various positions of the distribution mods, $DCFace_{all}$ introduces an alignment of these mods. Such behavior is coherent with the sensitivity study of the models:  the more the distribution of distances is equivalent across sub-groups the more stable the face recognition task will be across these sub-groups. 

Moreover, the quality of model embeddings can be evaluated by the diversity of different identities embeddings in the latent space. Thus DCFace and CASIA are clearly showing poorer embedding quality for underrepresented subgroups, whereas such a penalty for these subgroups is limited by $DCFace_{all}$ 

\section{Why analysing on \favcid}

\begin{figure}[ht]
    \centering
    \includegraphics[width=\linewidth]{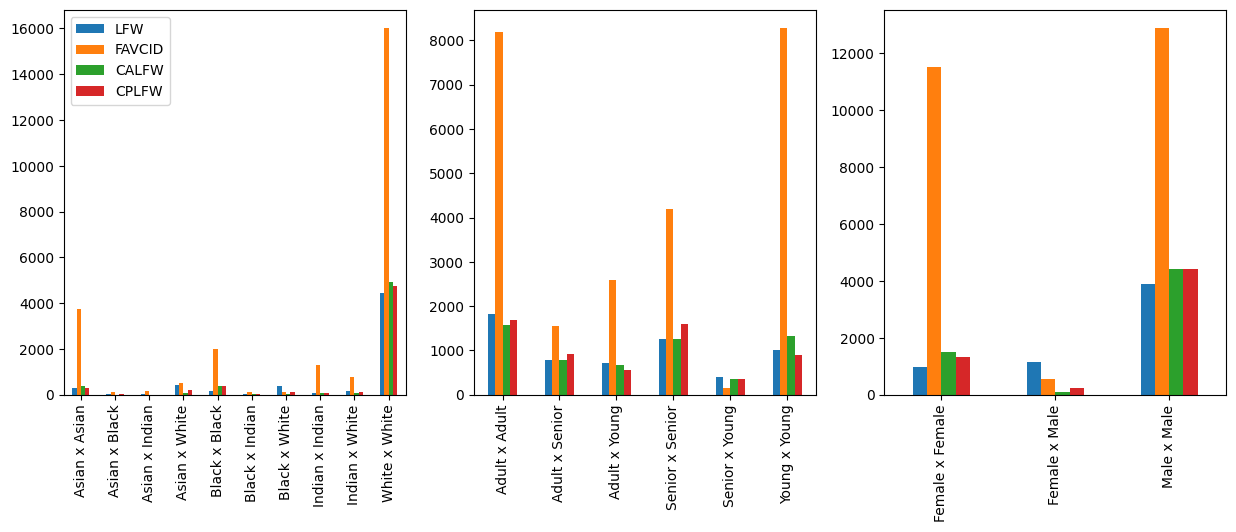}
    \caption{Pairs distribution for \favcid, \calfw , LFW and \cplfw}
    \label{fig_dataset_viz}
\end{figure}

As stated within the paper, we explicitly show in Figure \ref{fig_dataset_viz} the difference in the number of samples for LFW, \calfw, \cplfw and \favcid. The number of pairs between LFW and its variants (6000) is not suitable for a fairness study, since it introduces sparsity for some pairs subgroups and poor representativity for others. Such characteristics introduce singularity within the matrices used for regression by the former, and a critical increase of the study results uncertainty for the latter.

Moreover, for \calfw and \cplfw, we found that not only the number of samples is low (the same as in LFW),  but also that the models' performances are very sensitive to the alignment function used on the images. To maintain the performances, it is then required to use images provided by \cite{deng2019arcface} whose use is hard coded within the Adaface training pipeline \cite{kim2022adaface}. While showing performances coherent with the literature, these images do not have identity labels, which are required to infer ethnicity and gender consistently within identities. This post-treatment is necessary for fairness evaluation since FairFace presents for instance 15.6 percents of inconsistency for \favcid when predicting ethnicity and gender for images of the same person.

%
%

\end{document}